\documentclass[11pt,a4paper]{article}

\usepackage[a4paper,margin=1in]{geometry}
\usepackage[utf8]{inputenc}
\usepackage[T1]{fontenc}

\usepackage{cite}
\usepackage{url}
\usepackage{amsmath,amssymb,amsfonts}
\usepackage{algorithmic}
\usepackage{graphicx}
\usepackage{textcomp}
\usepackage{xcolor}
\usepackage{hyperref}
\usepackage{authblk}
\usepackage{array}
\usepackage{booktabs}

\hypersetup{
    colorlinks=true,
    linkcolor=blue,
    citecolor=blue,
    urlcolor=blue
}

\title{Beyond Fine-Tuning: Robust Food Entity Linking under Ontology Drift with FoodOntoRAG}

\author[1,2]{Jan Drole\\\small\href{https://orcid.org/0009-0002-3888-705X}{ORCID: 0009-0002-3888-705X}}
\author[2]{Ana Gjorgjevikj\\\small\href{https://orcid.org/0000-0002-5135-7718}{ORCID: 0000-0002-5135-7718}}
\author[2]{Barbara Korou\v{s}i\'c Seljak\\\small\href{https://orcid.org/0000-0001-7597-2590}{ORCID: 0000-0001-7597-2590}}
\author[2]{Tome Eftimov\\\small\href{https://orcid.org/0000-0001-7330-1902}{ORCID: 0000-0001-7330-1902}}

\affil[1]{Computer Systems Department, Jo\v{z}ef Stefan International Postgraduate School, Ljubljana, Slovenia}
\affil[2]{Computer Systems Department, Jo\v{z}ef Stefan Institute, Ljubljana, Slovenia}

\date{}

\begin{document}

\maketitle

\begin{center}
\small\textit{This manuscript is the authors' version of a paper published in the IEEE BigData 2025 proceedings. The final published version is available via IEEE Xplore.}
\end{center}

\vspace{0.5em}
\noindent\textbf{Funding.}
The authors acknowledge the support of the Slovenian Research Agency through program grant No. P2-0098 and project grant No. GC-0001. This work is also supported by the European Union under Grant Agreement 101211695 (HE MSCA-PF AutoLLMSelect). We also acknowledge the support of the EC/EuroHPC JU and the Slovenian Ministry of HESI via the project SLAIF (grant number 101254461).

\begin{abstract}
Standardizing food terms from product labels and menus into ontology concepts is a prerequisite for trustworthy dietary assessment and safety reporting.
The dominant approach to Named Entity Linking (NEL) in the food and nutrition domains fine-tunes Large Language Models (LLMs) on task-specific corpora. Although effective, fine-tuning incurs substantial computational cost, ties models to a particular ontology snapshot (i.e., version), and degrades under ontology drift. This paper presents FoodOntoRAG, a model- and ontology-agnostic pipeline that performs few-shot NEL by retrieving candidate entities from domain ontologies and conditioning an LLM on structured evidence (food labels, synonyms, definitions, and relations). A hybrid lexical--semantic retriever enumerates candidates; a selector agent chooses a best match with rationale; a separate scorer agent calibrates confidence; and, when confidence falls below a threshold, a synonym generator agent proposes reformulations to re-enter the loop. The pipeline approaches state-of-the-art accuracy while revealing gaps and inconsistencies in existing annotations. The design avoids fine-tuning, improves robustness to ontology evolution, and yields interpretable decisions through grounded justifications.
\end{abstract}

\vspace{0.5em}

\section{Introduction}
Food and nutrition data are dispersed across food product labels, online ingredient lists, and menu descriptions. Synonyms (e.g., \emph{icing sugar} vs.\ \emph{powdered sugar}), role-based phrasing (e.g., \emph{acidulant} vs.\ \emph{citric acid}), branding, and cultural and multilingual variants make downstream analytics brittle. Analysts and regulators require stable concept identifiers to (i) compute nutrient roll-ups, (ii) detect allergens and additives, and (iii) align reports across datasets and regions in a defensible, auditable way.

Integrating food composition and consumption data requires compliance with FAIR (Findable, Accessible, Interoperable, Reusable) principles~\cite{brinkley2025state, top2022Cultivating}, in line with EU strategies for sustainable agriculture and data sharing. Achieving interoperability depends on ontologies that unify vocabulary and semantics, such as FoodOn~\cite{dooley2018foodon}, SNOMED-CT~\cite{donnelly2006snomed}, Hansard~\cite{alexander2012hansard}, FoodEx2~\cite{european2015food}, and FoodOntoMap~\cite{popovski2019foodontomap}. Extraction based on Natural Language Processing (NLP) further enhances FAIRification via named-entity recognition (NER) and linking (NEL). The task is challenging due to long-tail terminology, overlapping synonymy, compositional spans in ingredient lists, and safety-sensitive use cases that demand calibrated decisions and principled abstention when no adequate match exists.

All artifacts and the curated Open Food Facts subset annotated with both FoodSEM and FoodOntoRAG are openly available: data on Zenodo\footnote{\url{https://zenodo.org/records/17347785}}, code on GitHub\footnote{\url{https://github.com/jan3657/onto_rag/tree/Restructured}}, and an interactive application for browsing the annotations\footnote{\url{https://jan3657-onto-rag-compare-app-restructured-atq5lz.streamlit.app/}}.

\textbf{Our contribution:} This paper presents \textit{FoodOntoRAG}, an ontology-aware retrieval-augmented pipeline for food NEL that combines hybrid lexical--semantic retrieval with an LLM-based selector guided by exact-match and specificity rules, a separate confidence scorer with abstention, and a synonym-based retry mechanism; decisions are grounded in structured FoodOn~\cite{dooley2018foodon} evidence (labels, synonyms, definitions, and relations) rather than task-specific fine-tuning. The accompanying artifacts comprise versioned ontology snapshots, pre-processing and indexing scripts (lexical and vector), prompt templates/configurations, and evaluation code to support end-to-end reproducibility. Empirically, the pipeline is first assessed on a public recipe-ingredient NEL benchmark and compared against a supervised state-of-the-art approach, demonstrating competitive accuracy. The study then evaluates a non-annotated corpus of branded product ingredient lists, again against the supervised baseline to quantify performance under real-world conditions.

\section{Background and Related Work}
\textbf{Food ontologies.} \textit{FoodOn} is an open-source ontology from Open Biological and Biomedical Ontologies (OBO) Foundry that describes entities with a ``food role'' across the farm-to-fork lifecycle, providing a structured vocabulary for sources, products, processes, qualities, and ingredients, with rich labels/synonyms and relations (e.g., \emph{has ingredient})~\cite{dooley2018foodon}. Organized into interoperable facets (e.g., organism parts, processing methods, quality attributes), FoodOn integrates with other life-science ontologies and is the primary target ontology in this work. Complementary resources are mentioned only for context: \textit{SNOMED CT} contributes clinically oriented food/allergen concepts useful for cross-walking to healthcare records~\cite{donnelly2006snomed}, and the \textit{Hansard} taxonomy (developed in the SAMUELS project) provides a ``Food and Drink'' semantic group that expands lexical coverage in public-text corpora~\cite{alexander2012hansard}.

\textbf{Entity normalization in food.} NEL maps textual mentions to ontology identifiers, enabling aggregation across heterogeneous sources despite synonyms and naming variation. Early food NEL studies included rule/lexicon pipelines and taxonomy aligners such as StandFood, which linked to FoodEx2 primarily via lexical similarity~\cite{eftimov2017standfood}. The release of food-annotated corpora (FoodBase~\cite{popovski2019foodbase}, CafeteriaFCD~\cite{ispirova2022cafeteriafcd}, CafeteriaSA~\cite{cenikj2022cafeteriasa}) catalyzed corpus-based methods: FoodNER fine-tuned BERT variants over multiple resources (Hansard, FoodOn, SNOMED CT) but showed limited cross-domain generalization (e.g., from recipes to scientific abstracts)~\cite{stojanov2021fine}, while SciFoodNER adapted training to scientific abstracts and linked to the same ontologies with mixed gains~\cite{cenikj2022scifoodner}. Recent zero-shot evaluations of general-purpose LLMs (e.g., GPT-3.5/4) report insufficient accuracy for precise ID linking. Recently, FoodSEM~\cite{gjorgjevikj2025foodsem}, a fine-tuned open-source LLM that sets a new state-of-the-art for food NEL to ontologies such as FoodOn, SNOMED-CT, and Hansard, achieves up to 98\% F1 and provides publicly available datasets, models, and benchmarks for advancing semantic understanding in the food domain. However, its NEL power is limited only to the entities presented in the training data. These observations---domain shift, ontology/version drift, and the cost of continuous fine-tuning---motivate approaches that minimize task-specific training. This work therefore adopts an ontology-aware Retrieval Augmented Generation (RAG) pipeline that retrieves candidates from FoodOn and delegates the final, evidence-grounded decision to an LLM.~\cite{lewis2020retrieval,ding2024ragSurvey,toro2024dynamic}

\textbf{Retrieval and RAG.} RAG couples a retriever (e.g.\ a dense index or knowledge-graph lookup) with a language model generator to ground outputs in external knowledge, improving factuality and decoupling knowledge updates from model weights, which allows refreshing knowledge without continual fine-tuning~\cite{lewis2020retrieval}. Foundational RAG systems demonstrate substantial gains on knowledge-intensive question answering by integrating dense retrieval with seq2seq generation~\cite{lewis2020retrieval}, and recent surveys distill design choices for indexing, retrieval, fusion and feedback~\cite{ding2024ragSurvey}. In biomedicine, entity-aware retrieval reduces token counts and inference latency while improving clinical information extraction~\cite{lopez2025clear}; map--reduce strategies such as BriefContext mitigate ``lost-in-the-middle'' effects when key information lies deep within long contexts~\cite{zhang2025longcontext}; knowledge-graph and hypergraph RAG frameworks improve accuracy and reduce hallucinations in medical question answering (QA)~\cite{wu2024medgrphrag, feng2025hyperrag}; and retrieval-augmented ontology generation methods like DRAGON-AI can generate new ontology terms and relationships with high precision, although expert curators remain essential~\cite{toro2024dynamic}. These results motivate our retrieval-first, ontology-aware pipeline for applications in the food and nutrition domains.

\begin{figure}[t]
    \centering
    \includegraphics[width=\textwidth]{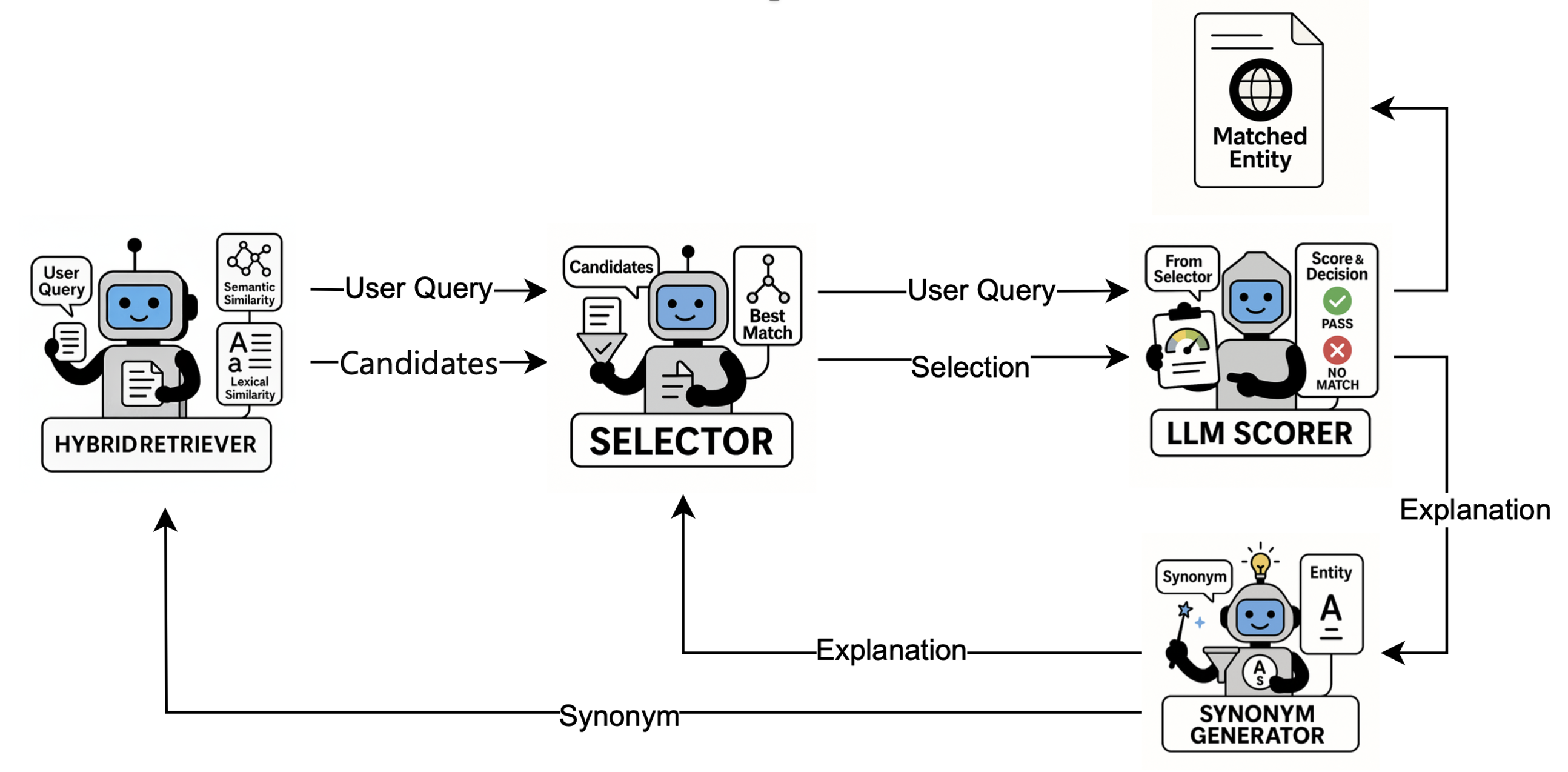}
    \caption{Visualization of the approach to named entity linking, composed of four different agents.}
    \label{fig:onto_rag_pipeline}
\end{figure}

\section{FoodOntoRAG methodology}
The approach used in this paper is visually presented in Fig.~\ref{fig:onto_rag_pipeline}. The pipeline is composed of four different agents, each contributing in a different way, and then looped in a feedback loop. In this section, each stage will be explained in detail with all the possible changes and adaptations. The first block of the pipeline is represented by a hybrid retriever agent, which is specialized for finding the most lexically similar entities to the user query as well as semantically similar matches. For the retriever to work correctly, the ontology data---comprising entity identifiers, preferred labels, synonyms, and hierarchical relationships from FoodOn---was first preprocessed. The selected entity with its corresponding user query is then passed to the selector agent, which in turn is responsible for selecting the single most suitable entity from the retrieved search space. In principle, as long as the underlying language model is multilingual, the input query can be provided in any language supported by the model; however, in this study, only English inputs were evaluated. That entity is selected based on its name and the set of synonyms provided by the ontology (e.g., FoodOn), as well as contextual similarity to the input mention and relations to other entities. Next in line is the LLM scorer agent, which is responsible for the decision whether the selection is a true match or a best match from the selector, which might have gotten a search space without the correct entity, or just made a mistake. There was a noticeable improvement separating the scorer from the selector during the development phase. The last part of the system feedback loop (and also where the feedback happens) is the synonym generator, which only gets activated in the case of scorer deciding the selected entity is not suitable. In that case, an explanation is also provided for why the selection is not correct. That explanation is then piped to the synonym generator, which generates another term for the user query with the same meaning, and the explanation is piped though the pipeline to the selector. Now that we have a synonym it is important to always compare the selection to the original user query.

\subsection{Data pre-processing}
Since the hybrid retrieving process works on two parts, the lexical and semantic similarity, the preprocessing is also done in two steps. The \textit{semantic similarity} search is done using the cosine distance of the entity to the ontology entities. For that, both entities from text and ontology need to be embedded using a text encoder model. For that a lightweight model is a good match and in this study the \textit{all-MiniLM-L6-v2}~\cite{allMiniLM_L6_v2} is selected for that task. The model was selected for its ability to represent text data well in a dense 384-dimensional vector space, as visible from its still competitive performance on certain general-purpose NLP tasks (e.g., information retrieval and semantic text similarity) in text embedding benchmarks such as MTEB\footnote{\url{https://huggingface.co/spaces/mteb/leaderboard}}, and for its fast inference due to the significantly smaller parameter size (22.7M parameters) compared to other state-of-the-art encoders. With that all the entities from the ontology are embedded and stored in a FAISS~\cite{douze2024faiss} vector database.

The \textit{lexical branch} of the hybrid retriever begins with ontology parsing that filters each Resource Description Framework (RDF) graph down to valid concept entities and collects their surface forms and relations. During parsing we restrict the candidate set to Web Ontology Language (OWL) classes and Simple Knowledge Organization System (SKOS) concepts while excluding properties and other non-term terms, ensuring every retained entity Uniform Resource Identifier (URI) can be converted to a CURIE for the corresponding ontology. For each valid entity we extract the preferred label, all synonym literals, textual definitions, hierarchical parents and ancestors, and human-readable relation names derived from the configured relation CURIEs. The extracted facets are merged per CURIE, validated against ontology-specific identifier patterns, and materialized as ontology-specific JSON dumps that provide the structured text payloads for indexing. These dumps are then transformed into Whoosh~\cite{Whoosh} lexical indexes whose schema stores labels, synonyms, and definitions while additionally indexing relation names as an auxiliary text field. Each ontology’s dump is iterated to populate that schema, producing a curated inverted index that powers keyword-style matching over the most salient lexical features of every concept.

\noindent\textbf{Example.} During preprocessing, each ontology concept is decomposed into its textual facets and relations before being serialized for retrieval. For instance, the FoodOn entity \texttt{FOODON:03302340} (``whole wheat flour'') is extracted from the RDF graph along with its preferred label, all declared synonyms (e.g., ``wholemeal flour,'' ``graham flour''), textual definition, and hierarchical relations such as \texttt{is\_a} \texttt{FOODON:00001210} (``wheat flour food product''). These components are merged under the same CURIE and validated to ensure identifier consistency. The resulting JSON representation for this entity contains key--value pairs such as:

\begin{verbatim}
{
  "curie": "FOODON:03302340",
  "label": "whole wheat flour",
  "synonyms": ["wholemeal flour", "graham flour"],
  "definition": "Undefined",
  "relations": {"is_a": ["FOODON:00001210"]}
}
\end{verbatim}

This structured record is then indexed by the lexical retriever (Whoosh)~\cite{Whoosh} under the fields \texttt{label}, \texttt{synonyms}, and \texttt{definition}, and simultaneously embedded as a dense vector for semantic retrieval. This preprocessing step ensures that both lexical and semantic retrievers operate over harmonized, ontology-consistent text representations.

\subsection{Hybrid retriever}
\label{sec:hybrid-retriever}
\textit{Goal.} Given a mention $m$ (optionally with local context $c$), return a small, high-recall set of FoodOn candidates for the selector.

\textit{Indexes.} The lexical index (Whoosh)~\cite{Whoosh} stores preferred labels, all synonyms, short definitions, and selected relation names from the preprocessed dump. The vector index (FAISS)~\cite{douze2024faiss} stores 384-dimensional embeddings (\textit{all-MiniLM-L6-v2}) for the same facets, one vector per concept.

\textit{Query.}
(i) \emph{Lexical:} BM25 over label/synonym-weighted fields; return top $K_{\text{lex}}{=}15$.
(ii) \emph{Semantic:} embed $m$ (and, if present, $c$ appended) and retrieve top $K_{\text{sem}}{=}15$ nearest neighbors from FAISS~\cite{douze2024faiss}.

\textit{Fusion and filtering.} Concatenate the lexical list followed by the semantic list; deduplicate by concept identifier; then apply light tie-breaks before truncation to $K_{\text{tot}}{=}K_{\text{lex}}{+}K_{\text{sem}}$ (default 30): (a) surface-form exact matches of $m$ to any label/synonym are promoted to the top; (b) candidates whose surface forms cover all tokens of $m$ are preferred; (c) otherwise, preserve within-branch order. No cross-branch score normalization is applied.

\textit{Output to selector.} For each candidate we provide the identifier (CURIE/IRI), preferred label, matched surface form (if any), top synonyms, a short definition snippet, and salient relations. These fields are inserted verbatim into the selector prompt.

This component is responsible for narrowing down the search space for the selector.
Since the selector uses model-agnostic prompting to an LLM, it operates within a limited context window---typically ranging from about 128{,}000 tokens for models such as GPT-4~\cite{openai2023gpt4} up to around 2 million tokens for models like Gemini 1.5 Pro~\cite{google2024gemini1_5}.
However, even this expanded capacity is often insufficient, as large ontologies or collections of ontologies can easily exceed several million tokens. Hence, the hybrid retriever serves a crucial role in reducing the search and reasoning space prior to selector invocation, ensuring that only the most contextually relevant segments are passed to the model within its manageable window, while maintaining efficiency and fidelity across the pipeline. The tunable parameters in this step include the number of retrieved entities from lexical search and semantic similarity search.

\subsection{Selector}
\label{sec:selector}
The selector is an instruction-following component that must choose a \emph{single} FoodOn term from the candidate list produced by retrieval. It is prompted with (i) the user entity, (ii) the enumerated candidate terms (ID, label, definition, synonyms), and (iii) a concise rubric. Two rules are encoded in the prompt template:
\begin{enumerate}
  \item \textbf{Exact-match preference.} Exact, case-insensitive matches to a candidate's label or synonym dominate partial or inferred matches.
  \item \textbf{Specificity rule.} When multiple candidates are plausible, prefer the most specific term over a broader hypernym.
\end{enumerate}

\paragraph*{Illustrative example.}
Given the user entity \texttt{LEBANESE} and the candidate set (abbrev.\ to the most relevant here):
\begin{itemize}
  \item \textbf{FOODON:03540141} --- \emph{01410 - pita bread (efsa foodex2)}; definition notes ``Lebanese bread'' as an alternative name.
  \item \textbf{FOODON:00005570} --- \emph{lebanon bologna}; a smoked, fermented beef sausage from Lebanon County, Pennsylvania.
  \item \textbf{FOODON:03302684} --- \emph{middle east bread}; broad category.
\end{itemize}
The selector applies the rules as follows:
\begin{enumerate}
  \item \emph{Exact-match preference.} While \texttt{LEBANESE} is not an exact label, the definition of \emph{pita bread} explicitly includes ``Lebanese bread'' as an alias, establishing a strong lexical-semantic link.
  \item \emph{Specificity rule.} Between \emph{pita bread} (a concrete food item) and \emph{middle east bread} (a broader class), the more specific \emph{pita bread} is preferred.
  \item \emph{Rejection of distractors.} \emph{lebanon bologna} is rejected despite surface string overlap (\texttt{Lebanon} vs.\ \texttt{Lebanese}); it denotes a Pennsylvania-style sausage unrelated to Lebanese cuisine.
\end{enumerate}

\noindent The selector then emits a compact JSON rationale (format-only illustrative):
\begin{verbatim}
{
  "chosen_id": "FOODON:03540141",
  "explanation": "The user entity 'LEBANESE'
    is used as a descriptor for 'Lebanese bread',
    which is explicitly listed as an
    alias for 'pita bread'"
}
\end{verbatim}

\noindent\textit{Note.} In this prompt, the selector chose the most suitable entity \emph{among the retrieved candidates}, yet the decision is still incorrect with respect to the user span because the true target is underspecified (\emph{Lebanese} as a demonym/cuisine) and may not correspond to a specific dish. This illustrates the need for a downstream confidence/verification agent that evaluates the selected ID against definition evidence and ambiguity cues, and can abstain or trigger refinement when the selection is not sufficiently justified.

\paragraph*{Output contract.}
The selector outputs JSON with only the chosen identifier and a brief justification. If no candidate qualifies under the rubric, it must return \texttt{"chosen\_id": "-1"} with an explanation of why no match is acceptable.

\paragraph*{Edge cases and guardrails.}
The prompt emphasizes conservative behavior on (i) geographic adjectives (\emph{Lebanese}, \emph{Syrian}) that may denote cuisines, dishes, or demonyms; (ii) homonyms (e.g., \emph{bologna} vs.\ city/cold cut); and (iii) regulated substances where label/number mismatches are disqualifying. When ambiguity remains among near-peers, the selector favors the candidate whose definition/synonyms most directly substantiate the link in the user span, deferring broader categories unless no specific item qualifies.

\subsection{LLM Scorer}
A separate assessor evaluates the selector’s proposal. It returns a continuous confidence score in the interval $[0,1]$ and a concise justification grounded in a rubric that (i) heavily penalizes identity mismatches between distinct substances, and (ii) treats formulation indicators (e.g., \emph{salt}, \emph{hydrate}, \emph{lake}) as non-identity-altering unless explicitly stated. If the score is $<\tau$, the scorer also suggests up to three alternative candidates from the retrieved list and trigger retries.

\textbf{Example of scorer's explanation}, that showcases additional reasoning over just an isolated selector:
\textit{``Poor Match. The user entity 'LEBANESE' refers to a nationality or origin, while the chosen ontology term 'pita bread' refers to a type of food. Although 'Lebanese bread' is listed as a synonym for pita bread, the primary entity is not the bread itself but the origin. Rule 2 is implicitly applied as the entities are fundamentally different categories (nationality vs.\ food item).''}

\subsection{Synonym Generator (Feedback Loop)}
\label{sec:synonym-generator}

\paragraph*{Role.}
When invoked by the confidence scorer (see \S\textit{LLM Scorer}), this component proposes \emph{query reformulations} for the original mention so that a second retrieval--selection pass can surface missed ontology entries. It deliberately avoids re-ranking the same candidate list.

\paragraph*{Decoupled design.}
The generator is a dedicated agent, separate from both selector and scorer prompts. This separation reduced prompt echoing and prior-selection anchoring, yielding more diverse and semantically precise alternatives across registers (commodity vs.\ scientific).

\paragraph*{Interface.}
Inputs: (i) the original user mention $q$ (optionally with minimal local context), and (ii) a concise \texttt{failure\_reason} string provided by the scorer (e.g., identity synonyms mismatch, processing-state mismatch, ambiguity).
Output: a JSON object with up to five strings:
\begin{verbatim}
{
  "synonyms": ["<alt1>", "<alt2>", "..."]
}
\end{verbatim}
Constraints: exclude the original surface form $q$; remain ontology-agnostic (no hard-facet assumptions); prefer high-precision paraphrases over broad hypernyms.

\paragraph*{Prompt contract (abridged).}
The agent is instructed to act as a query-expansion specialist for food/scientific domains and to produce \emph{up to five} alternatives that:
(i) include direct synonyms (e.g., ``baking soda'' for ``sodium bicarbonate''),
(ii) offer phrasing variants (e.g., inverted qualifiers like ``sugar, powdered''),
(iii) introduce scientific/technical names when applicable (e.g., ``ascorbic acid'' for ``vitamin C''), and
(iv) add common/lay terms.
The response must be \emph{JSON only} under a single \texttt{"synonyms"} key, with no surrounding prose.

\paragraph*{Failure-aware conditioning.}
The \texttt{failure\_reason} focuses generation on the observed error mode. For an \textit{identity mismatch}, near-synonyms are prioritized over hypernyms; for a \textit{processing-state mismatch}, state-specific variants are proposed (e.g., pasteurized, dried, powdered). This lightweight conditioning increases the likelihood that the second pass retrieves the correct concept.

\paragraph*{One-hop loop and safeguards.}
Upon invocation:
\begin{enumerate}
  \item Generate $\tilde{Q}$ with at most five reformulations.
  \item Augment the query set with $\tilde{Q}$ and re-run hybrid retrieval with unchanged parameters; invoke the selector and scorer.
  \item If confidence remains insufficient, stop (no further synonym hops) and surface up to three alternatives with explanations for review.
\end{enumerate}
All re-selected candidates are checked against both the reformulated mention and the original $q$ to prevent semantic drift.

\paragraph*{Implementation notes and limitations.}
The generator is stateless and does not consume ontology facets directly. It currently lacks explicit diversity control or calibration of expansion quality; risk is mediated by the downstream scorer and the one-hop cap. Logs capture prompt versions and outputs for audit and error analysis.

\subsection{Output Serialization}
For each mention, the system emits JSON containing: the FoodOn identifier (CURIE/IRI), preferred label, the selector’s rationale, the scorer's rationale, the scorer’s confidence, and the original mention. When confidence $<\tau$, the output additionally includes the scorer’s rejection rationale and synonym proposals to support human review.

\section{Results}
\label{sec:results}
This section first reports the entity-linking performance of FoodOntoRAG on the CafeteriaFCD corpus. The \textbf{CafeteriaFCD} corpus~\cite{ispirova2022cafeteriafcd} is a curated dataset of 1,000 annotated recipes linking food entities to Hansard, FoodOn, and SNOMED-CT. Created using the NCBO Annotator and refined by experts, it contains a total of 7,429 ingredient entity annotations of which 948 are unique ingredient names. To avoid redundancy and focus on generalization, the evaluation was conducted on these 948 unique mentions. Next, we further evaluated FoodOntoRAG on ingredient lists from branded food products, which typically lack quantity information unlike our recipe-based training data. In this evaluation, we also compared the results with FoodSEM, the fine-tuned supervised model; however, no comparison was made on CafeteriaFCD, as FoodSEM was trained on data that include its instances. The evaluation used 119 ingredients from six branded products obtained via the Open Food Facts API\footnote{\url{https://world.openfoodfacts.org/}}, each providing product name, ingredients, nutrient values, and food group. The sample size is limited because all mappings were manually verified by two domain experts using a custom evaluation application. This web-based tool displays, side by side, the entity-linking outputs produced by FoodSEM and FoodOntoRAG for the same text span, along with links to the corresponding ontology entries and the initially matched entity. Experts can then select which system provides the better match, indicate if both are correct, or flag both as incorrect. The interface thus enables transparent comparison, error inspection, and consensus-based validation of model outputs. Finally, we present the user interface of this evaluation app as part of the qualitative analysis supporting domain expert review.

\subsection{CafeteriaFCD evaluation}

\begin{table}[t]
  \centering
  \caption{Acc@1 versus confidence threshold $\tau$ on CafeteriaFCD ($n{=}948$) with Gemini 2.5 Flash Lite; $K_{\text{lex}}{=}15$, $K_{\text{sem}}{=}15$, concatenation fusion. ``First'' is the initial attempt; ``Final'' is after any retry/synonyms if triggered.}
  \label{tab:acc_vs_tau}
  \resizebox{\textwidth}{!}{%
  \begin{tabular}{lrrrrrr}
    \toprule
    $\tau$ & Acc@1 (overall) & Acc@1 first & Acc@1 final & Retry rate & Syn.\ rate \\
    \midrule
    0.4 & 59.8\% & 59.7\% & 59.2\% & 7.5\% & 3.8\% \\
    0.5 & 58.8\% & 59.9\% & 58.1\% & 9.3\% & 3.7\% \\
    0.6 & 57.5\% & 60.0\% & 58.4\% & 8.4\% & 4.1\% \\
    0.7 & 58.6\% & 60.2\% & 58.6\% & 9.9\% & 4.3\% \\
    0.8 & 57.2\% & 59.6\% & 57.2\% & 11.7\% & 5.8\% \\
    \bottomrule
  \end{tabular}%
  }
\end{table}

The evaluation results summarized in Table~\ref{tab:acc_vs_tau} present the performance of FoodOntoRAG on the CafeteriaFCD corpus under different confidence thresholds ($\tau$) applied by the scorer to determine whether a selected entity should be accepted or retried. The reported metrics include the final top-1 accuracy (correct FoodOn match) after the full pipeline, including retries if triggered (Acc@1 overall), the accuracy on the first attempt before any synonym-based retries (Acc@1 first), the accuracy after all retry/synonym loops complete (Acc@1 final), as well as the retry rate that is the proportion of mentions for which the scorer rejected the initial selection, and the synonym rate, indicating the share of mentions where the synonym generator proposed an alternative query.

Formally, the metrics are defined as:
\begin{itemize}
  \item \(i\): index of a mention in the corpus; \(M\) is the total number of evaluated mentions.
  \item \(T_i\): the set of gold-acceptable FoodOn targets for mention \(i\).
  \item \(y_i^{\text{first}}\): top-1 prediction from the \emph{first} pass (before any retries/synonyms).
  \item \(y_i^{\text{final}}\): top-1 prediction after the \emph{full pipeline} completes (including retries/synonym loops). If the pipeline fails to return a result, set \(y_i^{\text{final}}=\varnothing\).
  \item \(H_i\): number of attempts/hops for mention \(i\) (a retry occurred iff \(H_i>1\)).
  \item \(S_i \in \{0,1\}\): indicator that at least one synonym-generated query was used for mention \(i\).
  \item \(\mathbf{1}\{\cdot\}\): indicator function (1 if the condition holds, else 0).
\end{itemize}

The metrics are computed over all \(M\) mentions (failures count as incorrect for accuracies):

\begin{align}
\text{Acc@1\ (overall)}
&=\frac{1}{M}\sum_{i=1}^{M}\mathbf{1}\!\left(\, \big[y_i^{\text{first}}\in T_i\big]\;\lor\;\big[y_i^{\text{final}}\in T_i\big] \,\right), \\[6pt]
\text{Acc@1\ first}
&=\frac{1}{M}\sum_{i=1}^{M}\mathbf{1}\!\left(y_i^{\text{first}}\in T_i\right), \\[6pt]
\text{Acc@1\ final}
&=\frac{1}{M}\sum_{i=1}^{M}\mathbf{1}\!\left(y_i^{\text{final}}\in T_i\right), \\[6pt]
\mathrm{Retry\ rate}
&=\frac{1}{M}\sum_{i=1}^{M}\mathbf{1}\!\left(H_i>1\right), \\[6pt]
\mathrm{Synonym\ rate}
&=\frac{1}{M}\sum_{i=1}^{M} S_i \, .
\end{align}

Across all thresholds ($\tau$), the system maintains a stable accuracy between 57\% and 60\%, demonstrating the robustness of the retrieval-augmented approach without fine-tuning. The difference between the first and final accuracies remains within one to two percentage points, suggesting that the majority of correct mappings are achieved on the initial attempt, while synonym-based retries contribute only modest improvements. As the threshold $\tau$ increases, the system becomes more conservative, leading to higher retry and synonym rates (from 7.5\% to 11.7\%), yet without a proportional gain in accuracy. This behavior indicates diminishing returns from overly strict confidence filtering.

Overall, the best trade-off between accuracy and computational cost is observed around $\tau = 0.6\text{--}0.7$, where FoodOntoRAG achieves approximately 58--59\% accuracy while keeping retry operations manageable. These results confirm that the confidence-controlled feedback loop---comprising the scorer and synonym generator---enhances interpretability and error calibration, while preserving competitive performance under ontology drift, thus validating the efficiency and robustness of the proposed few-shot RAG-based entity linking pipeline.

Upon examining the results, we observed that some annotations generated by FoodOntoRAG, initially marked as false during evaluation due to mismatches with CafeteriaFCD annotations, were, in fact, correct. This discrepancy arises because certain food entities can correspond to multiple valid CURIEs depending on the ontology hierarchy level, which may differ from the branch level used as a guideline in CafeteriaFCD annotation. Examples of such cases are shown in Table~\ref{tab:adjudication-cases}, where FoodOntoRAG annotations are compared with those from the reference corpus. For instance, ``onion'' can correspond to both a general ingredient class (FoodOn:03316347) and a biological taxon (NCBITaxon:4679). Similar cases appear for walnuts, artichoke, parmesan cheese, and soy sauce. These examples illustrate that FoodOntoRAG often retrieves semantically correct but hierarchically different entities, highlighting the challenge of aligning evaluation corpora with ontology granularity and the system’s capability to capture valid alternative representations. After identifying these mismatches, we manually re-evaluated them to account for such hierarchical variations and recalculated the final accuracy, which is now 97\%.

\begin{table}[t]
  \centering
  \caption{Representative adjudicated cases illustrating unannotated-but-correct predictions.}
  \label{tab:adjudication-cases}
  \scriptsize
  \begin{tabular}{p{0.21\linewidth}p{0.34\linewidth}p{0.34\linewidth}}
    \toprule
    Mention & Model decision (label; CURIE) & Dataset annotations (label; CURIE) \\
    \midrule
    ONION & onion; FOODON:03316347 & onion (raw); FOODON:03301704; Allium \textit{cepa}; NCBITaxon:4679 \\
    \midrule
    WALNUTS & walnuts (FoodEx2); FOODON:03541166 & \textit{Juglans}; NCBITaxon:16718 \\
    \midrule
    ARTICHOKE & artichoke head; FOODON:00003573 & \textit{Cynara cardunculus} var.\ \textit{scolymus}; NCBITaxon:59895 \\
    \midrule
    PARMESAN CHEESE & parmesan cheese (block); FOODON:03303675 & parmesan cheese food product; FOODON:00003247 \\
    \midrule
    SOY SAUCE & soy sauce (FoodEx2); FOODON:03544111 & soy sauce; FOODON:03301115 \\
    \midrule
    TORTILLA & tortilla (FoodEx2); FOODON:03540143 & tortilla; FOODON:03307668 \\
    \midrule
    PROVOLONE & cheese, provolone (FoodEx2); FOODON:03542844 & provolone cheese; FOODON:03302916 \\
    \midrule
    GARBANZO & garbanzo bean (whole); FOODON:00002774 & \textit{Cicer arietinum}; NCBITaxon:3827 \\
    \midrule
    HONEY & honey (FoodEx2); FOODON:03543011 & honey; UBERON:0036016 \\
    \midrule
    ROSEMARY & rosemary (FoodEx2); FOODON:03540863 & rosemary food product; FOODON:00002212 \\
    \bottomrule
  \end{tabular}
\end{table}

\begin{figure}[t]
    \centering
    \includegraphics[width=\textwidth]{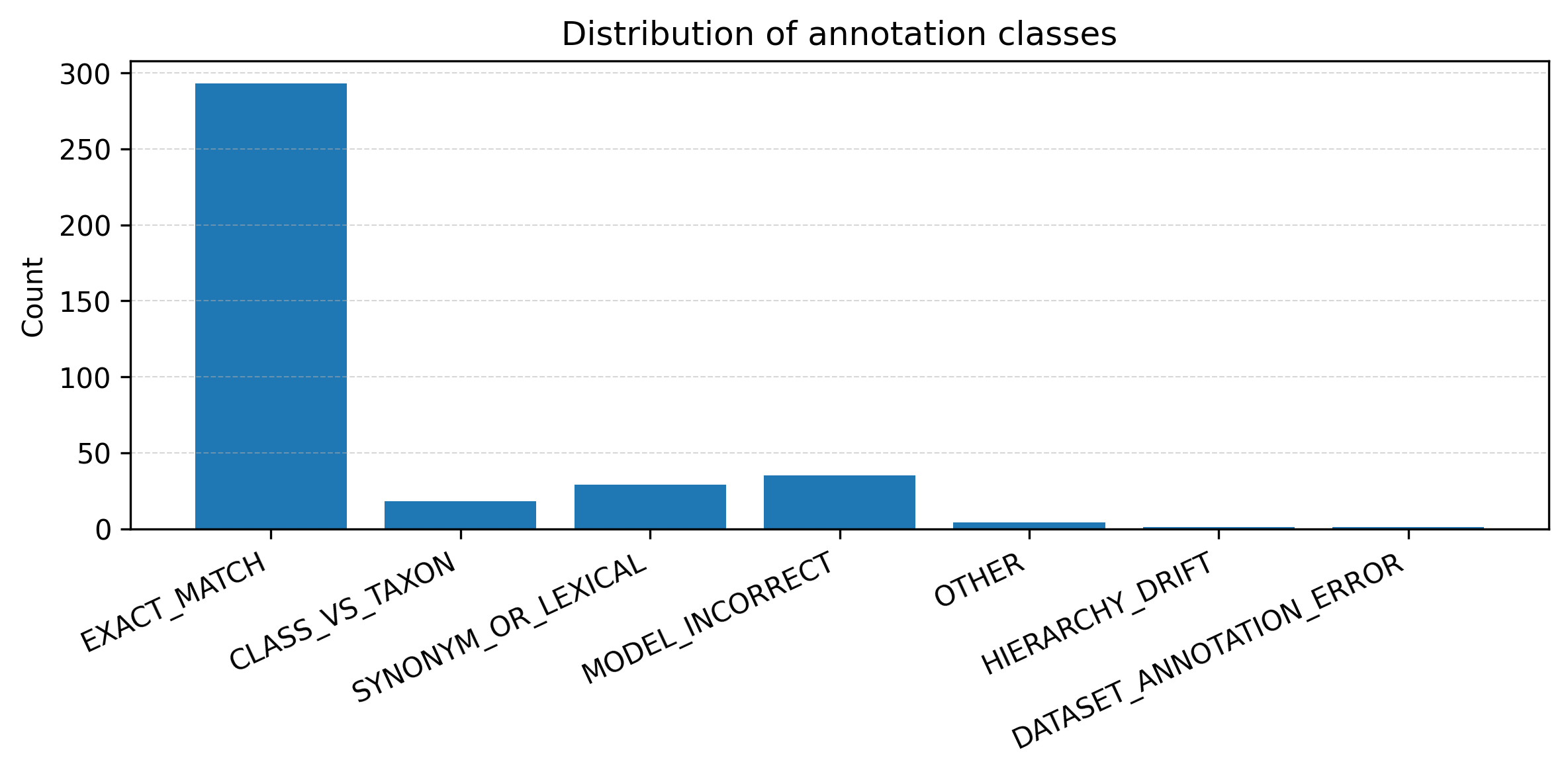}
    \caption{Distribution of annotation classes and disagreement types after ontology-aware adjudication of CafeteriaFCD mismatches.}
    \label{fig:drift_breakdown}
\end{figure}

\subsection{Ontology drift analysis of CafeteriaFCD mismatches}

To systematically extend the qualitative insights from Table~\ref{tab:adjudication-cases}, we automatically adjudicated all CafeteriaFCD cases where FoodOntoRAG's prediction disagreed with the reference annotations. For each such mention, we constructed a minimal JSON record containing only the original query, the FoodOntoRAG-chosen FoodOn term, and the set of ground-truth ontology terms. A constrained Gemini-based annotator was then prompted to (i) select the most appropriate gold term for the query and (ii) classify the relation between the chosen term and this selected gold term into one of several ontology-aware categories: \textsc{Exact\_Match}, \textsc{Class\_vs\_Taxon}, \textsc{Hierarchy\_Drift}, \textsc{Synonym\_or\_Lexical}, \textsc{Cross\_Ontology\_Equivalent}, \textsc{Dataset\_Annotation\_Error}, \textsc{Model\_Incorrect}, or \textsc{Other}. The prompt enforced a JSON-only response and deterministic rubric-based decisions to support transparent post-hoc analysis.

Intuitively, these buckets capture distinct modes of disagreement:
\begin{itemize}
    \item \textsc{Exact\_Match}: The chosen term is equivalent to the selected gold term (identical CURIE or semantically indistinguishable under the corpus guidelines). \emph{Example:} ``PARMESAN CHEESE'' mapped to a specific parmesan block term versus a generic \emph{parmesan cheese food product}.
    \item \textsc{Class\_vs\_Taxon}: One side refers to a food product or ingredient class, the other to a biological taxon describing the source organism. \emph{Example:} ``WALNUTS'' mapped to a FoodOn walnuts product concept versus the genus \emph{Juglans} (NCBITaxon).
    \item \textsc{Synonym\_or\_Lexical}: The terms denote the same concept but differ only in wording, spelling, or synonym choice, without a meaningful ontological or hierarchical shift.
    \item \textsc{Model\_Incorrect}: FoodOntoRAG's chosen term does not match the query's meaning, even though a suitable gold term is available.
    \item \textsc{Other}: Residual ambiguous or mixed cases that do not reliably fit any of the above categories.
    \item \textsc{Hierarchy\_Drift}: Both terms belong to the same ontology but differ in hierarchy level (parent$\rightarrow$child, child$\rightarrow$parent, or close siblings), reflecting granularity rather than substance change.
    \item \textsc{Dataset\_Annotation\_Error}: The selected gold term is clearly inconsistent with the mention (e.g., wrong ingredient or miscoded identifier), indicating an error in the original annotation.
\end{itemize}

The resulting distribution of adjudicated labels is summarized in Fig.~\ref{fig:drift_breakdown}. Out of 381 initially flagged ``errors,'' 293 (76.9\%) were reclassified as \textsc{Exact\_Match}, indicating that FoodOntoRAG's prediction is semantically aligned with the selected gold term despite minor representational differences. A further 29 cases (7.6\%) were labeled \textsc{Synonym\_or\_Lexical}, where discrepancies stem solely from alternative surface forms, and 18 cases (4.7\%) as \textsc{Class\_vs\_Taxon}, reflecting systematic substitutions between ingredient/product classes and their corresponding biological taxa. Only 35 instances (9.2\%) were assigned to \textsc{Model\_Incorrect}, capturing clear semantic mismatches, while \textsc{Hierarchy\_Drift}, \textsc{Dataset\_Annotation\_Error}, \textsc{Cross\_Ontology\_Equivalent}, and \textsc{Other} together accounted for fewer than 2\% of cases.

These findings show that the majority of apparent mismatches are explained by ontology drift, granularity choices, or synonymous representations rather than genuine model failures. In other words, FoodOntoRAG frequently proposes ontology-consistent alternatives that fall outside the single-CURIE assumptions of the original CafeteriaFCD guidelines, underscoring both the limitations of strict exact-match evaluation and the value of our ontology-aware analysis.

\subsection{OpenFoodFacts evaluation}
Here, we report only the final Acc@1 results for FoodOntoRAG and FoodSEM. The evaluation involved 119 ingredients from six branded products retrieved via the Open Food Facts API, each including product name, ingredients, nutrient values, and food group. The dataset size is limited as all mappings were manually validated by three domain experts using a custom-built evaluation app, whose interface is shown in the next section. FoodOntoRAG achieved an accuracy of 90.7\%, while FoodSEM reached 36.9\%. It is important to note that FoodSEM was trained on CURIEs from the CafeteriaFCD corpus, which primarily includes food ingredients. In contrast, the Open Food Facts data also contain colorants and chemical additives---entities unseen during FoodSEM training---explaining the observed performance gap.
A second annotator, using the same evaluation app on the same six products, obtained Acc@1 of 83.3\% for FoodOntoRAG and 29.2\% for FoodSEM, indicating a slight variation dependent on the annotator while confirming the substantial performance gap.

\subsection{Customized Application for Expert Validation of Evaluation Results}

The Ontology Mapping Comparator is an interactive tool developed to compare ontology mappings generated by two approaches---FoodSem and OntoRAG---for the same set of entities. It integrates results from both systems and enriches them with definitions and synonyms retrieved from an ontology dump, allowing for a transparent and systematic comparison of mappings.

\begin{figure}[t]
    \centering
    \includegraphics[width=0.9\textwidth]{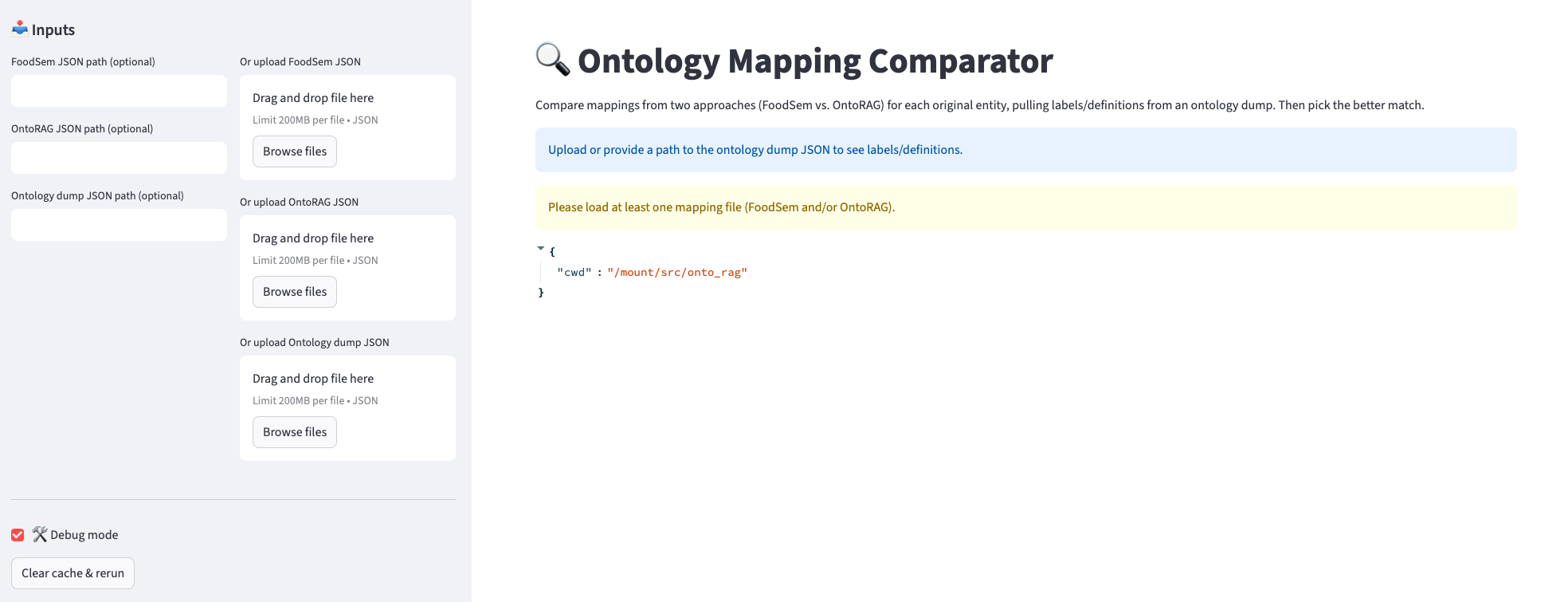}
    \caption{The Ontology Mapping Comparator input user interface.}
    \label{fig:input_uI}
\end{figure}

The interface consists of three main input components: (1) a JSON file containing FoodSem mappings, (2) a JSON file containing OntoRAG mappings, and (3) a JSON ontology dump providing semantic information such as labels, definitions, and synonyms for the referenced ontology terms (see Fig.~\ref{fig:input_uI}). Once the files are uploaded, the tool automatically aligns corresponding entities from both mapping systems.

\begin{figure}[t]
    \centering
    \includegraphics[width=0.9\textwidth]{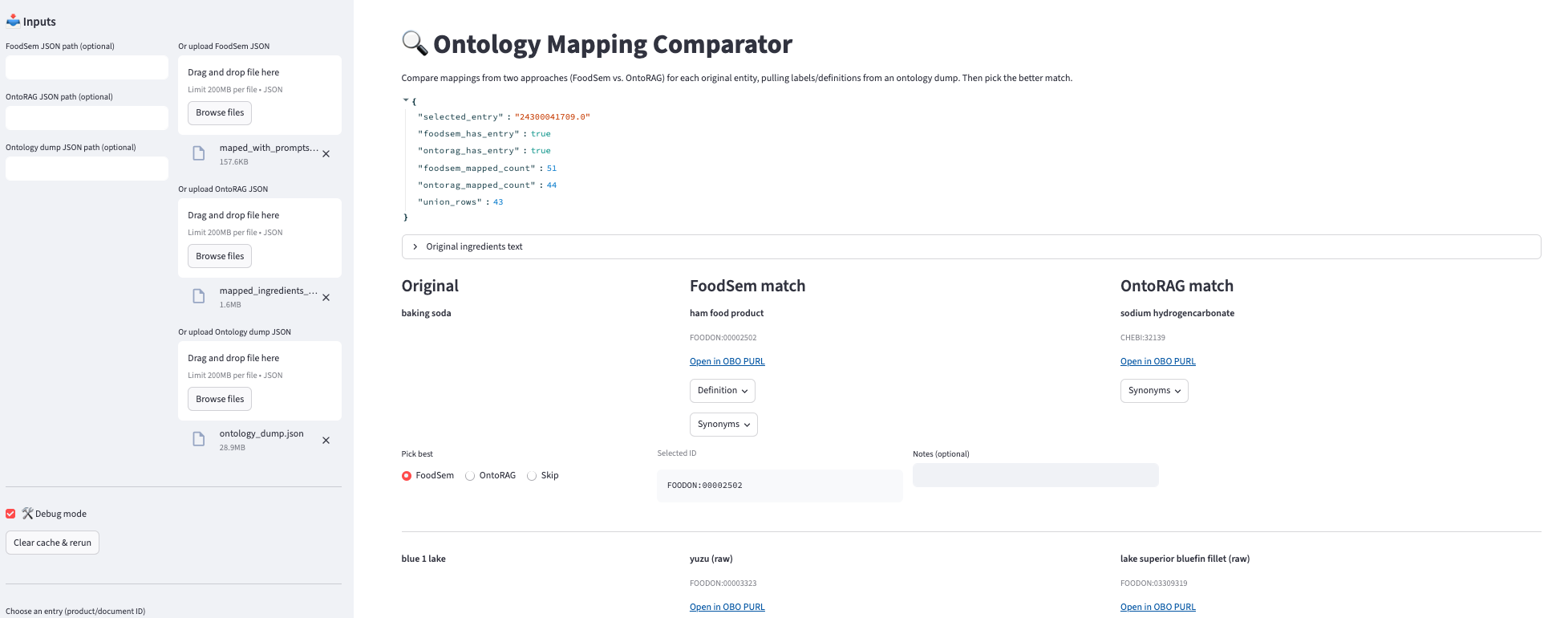}
    \caption{The Ontology Mapping Comparator result user interface.}
    \label{fig:result_uI}
\end{figure}

For each original ingredient, the interface displays the entity name, followed by two matched candidates---one from FoodSem and one from OntoRAG---alongside their ontology identifiers, definitions, and synonyms (see Fig.~\ref{fig:result_uI}). The user can directly inspect these mappings, access the corresponding ontology entries via OBO PURLs, and decide which mapping better represents the original entity. Additional statistics summarizing the number of mapped entries and overlaps between systems are also provided.

By offering a visual and interactive environment for manual validation, the Ontology Mapping Comparator supports the evaluation of automated ontology alignment methods, ensuring that the final mappings are both semantically accurate and explainable.

\section{Discussion}
The main strength of FoodOntoRAG lies in its model- and ontology-agnostic design, which achieves robust entity linking without fine-tuning. By relying on retrieval-augmented generation, the approach remains resilient to ontology drift and can easily adapt to evolving vocabularies. The combination of hybrid lexical--semantic retrieval and confidence-based feedback enables stable accuracy across datasets while ensuring interpretability through explicit rationales. Moreover, the pipeline’s modular structure allows rapid extension to new ontologies or domains without additional model training.

However, several limitations suggest directions for future work. First, the selection of text embedding models significantly affects retrieval quality; exploring larger or domain-specific embedding models may enhance semantic coverage. Second, the system’s sensitivity to the number of retrieved entities in both lexical and semantic searches influences precision and latency, requiring systematic tuning for scalability. Third, while the current evaluation focuses on food entities, broader applicability will require integrating additional domains such as chemicals, drugs, and diseases, along with their corresponding ontologies (e.g., ChEBI, DrugBank, DOID). Finally, further improvement could involve dynamic ontology fusion and adaptive retriever weighting, enabling balanced performance across heterogeneous knowledge sources.

\section{Conclusion}
This study introduced FoodOntoRAG, a few-shot, ontology-agnostic pipeline for food entity linking that bridges lexical and semantic retrieval with LLM-based reasoning. The approach achieves competitive accuracy without fine-tuning, demonstrating adaptability to evolving ontologies and transparency in decision-making. The results confirm the feasibility of retrieval-augmented linking as a sustainable alternative to supervised fine-tuned models, paving the way for scalable, cross-domain semantic integration in food, nutrition, and related life-science applications.

\bibliographystyle{IEEEtran}

\begin{thebibliography}{10}
\providecommand{\url}[1]{#1}
\csname url@samestyle\endcsname
\providecommand{\newblock}{\relax}
\providecommand{\bibinfo}[2]{#2}
\providecommand{\BIBentrySTDinterwordspacing}{\spaceskip=0pt\relax}
\providecommand{\BIBentryALTinterwordstretchfactor}{4}
\providecommand{\BIBentryALTinterwordspacing}{\spaceskip=\fontdimen2\font plus
\BIBentryALTinterwordstretchfactor\fontdimen3\font minus \fontdimen4\font\relax}
\providecommand{\BIBforeignlanguage}[2]{{%
\expandafter\ifx\csname l@#1\endcsname\relax
\typeout{** WARNING: IEEEtran.bst: No hyphenation pattern has been}%
\typeout{** loaded for the language `#1'. Using the pattern for}%
\typeout{** the default language instead.}%
\else
\language=\csname l@#1\endcsname
\fi
#2}}
\providecommand{\BIBdecl}{\relax}
\BIBdecl

\bibitem{brinkley2025state}
S.~Brinkley, J.~J. Gallo-Franco, N.~V{\'a}zquez-Manjarrez, J.~Chaura, N.~K. Quartey, S.~B. Toulabi, M.~T. Odenkirk, E.~Jermendi, M.-A. Laporte, H.~E. Lutterodt \emph{et~al.}, ``The state of food composition databases: data attributes and fair data harmonization in the era of digital innovation,'' \emph{Frontiers in Nutrition}, vol.~12, p. 1552367, 2025.

\bibitem{top2022Cultivating}
J.~Top, S.~Janssen, H.~Boogaard, R.~Knapen, and G.~{\c{S}}im{\c{s}}ek-{\c{S}}enel, ``Cultivating fair principles for agri-food data,'' \emph{Computers and Electronics in Agriculture}, vol. 196, p. 106909, 2022.

\bibitem{dooley2018foodon}
D.~M. Dooley, E.~J. Griffiths, G.~S. Gosal, P.~L. Buttigieg, R.~Hoehndorf, M.~C. Lange, L.~M. Schriml, F.~S. Brinkman, and W.~W. Hsiao, ``Foodon: a harmonized food ontology to increase global food traceability, quality control and data integration,'' \emph{npj Science of Food}, vol.~2, no.~1, p.~23, 2018.

\bibitem{donnelly2006snomed}
K.~Donnelly \emph{et~al.}, ``Snomed-ct: The advanced terminology and coding system for ehealth,'' \emph{Studies in health technology and informatics}, vol. 121, p. 279, 2006.

\bibitem{alexander2012hansard}
M.~Alexander and J.~Anderson, ``The hansard corpus, 1803-2003,'' 2012.

\bibitem{european2015food}
E.~F. S.~A. (EFSA), ``The food classification and description system foodex 2 (revision 2),'' Wiley Online Library, Tech. Rep., 2015.

\bibitem{popovski2019foodontomap}
G.~Popovski, B.~Korousic-Seljak, and T.~Eftimov, ``Foodontomap: Linking food concepts across different food ontologies.'' in \emph{KEOD}, 2019, pp. 195--202.

\bibitem{eftimov2017standfood}
T.~Eftimov, P.~Koro{\v{s}}ec, and B.~Korou{\v{s}}i{\'c}~Seljak, ``Standfood: standardization of foods using a semi-automatic system for classifying and describing foods according to foodex2,'' \emph{Nutrients}, vol.~9, no.~6, p. 542, 2017.

\bibitem{popovski2019foodbase}
G.~Popovski, B.~K. Seljak, and T.~Eftimov, ``Foodbase corpus: a new resource of annotated food entities,'' \emph{Database}, vol. 2019, p. baz121, 2019.

\bibitem{ispirova2022cafeteriafcd}
G.~Ispirova, G.~Cenikj, M.~Ogrinc, E.~Valen{\v{c}}i{\v{c}}, R.~Stojanov, P.~Koro{\v{s}}ec, E.~Cavalli, B.~Korou{\v{s}}i{\'c}~Seljak, and T.~Eftimov, ``Cafeteriafcd corpus: food consumption data annotated with regard to different food semantic resources,'' \emph{Foods}, vol.~11, no.~17, p. 2684, 2022.

\bibitem{cenikj2022cafeteriasa}
G.~Cenikj, E.~Valen{\v{c}}i{\v{c}}, G.~Ispirova, M.~Ogrinc, R.~Stojanov, P.~Koro{\v{s}}ec, E.~Cavalli, B.~K. Seljak, and T.~Eftimov, ``Cafeteriasa corpus: scientific abstracts annotated across different food semantic resources,'' \emph{Database}, vol. 2022, p. baac107, 2022.

\bibitem{stojanov2021fine}
R.~Stojanov, G.~Popovski, G.~Cenikj, B.~Korou{\v{s}}i{\'c}~Seljak, and T.~Eftimov, ``A fine-tuned bidirectional encoder representations from transformers model for food named-entity recognition: Algorithm development and validation,'' \emph{Journal of medical Internet research}, vol.~23, no.~8, p. e28229, 2021.

\bibitem{cenikj2022scifoodner}
G.~Cenikj, G.~Petelin, B.~K. Seljak, and T.~Eftimov, ``Scifoodner: food named entity recognition for scientific text,'' in \emph{2022 IEEE International Conference on Big Data (Big Data)}.\hskip 1em plus 0.5em minus 0.4em\relax IEEE, 2022, pp. 4065--4073.

\bibitem{gjorgjevikj2025foodsem}
A.~Gjorgjevikj, M.~Martinc, G.~Cenikj, S.~D{\v{z}}eroski, B.~Korou{\v{s}}i{\'c}~Seljak, and T.~Eftimov, ``Foodsem: Large language model specialized in food named-entity linking,'' in \emph{International Conference on Discovery Science}.\hskip 1em plus 0.5em minus 0.4em\relax Springer, 2025, pp. 395--410.

\bibitem{lewis2020retrieval}
P.~Lewis, E.~Perez, A.~Piktus, F.~Petroni, V.~Karpukhin, N.~Goyal, H.~K{\"u}ttler, M.~Lewis, W.-t. Yih, T.~Rockt{\"a}schel \emph{et~al.}, ``Retrieval-augmented generation for knowledge-intensive nlp tasks,'' \emph{Advances in neural information processing systems}, vol.~33, pp. 9459--9474, 2020.

\bibitem{ding2024ragSurvey}
\BIBentryALTinterwordspacing
Y.~Ding, W.~Fan, L.-B. Ning, S.~Wang, H.~Li, D.~Yin, T.-S. Chua, and Q.~Li, ``A survey on rag meets {LLMs}: Towards retrieval-augmented large language models,'' \emph{arXiv preprint arXiv:2405.06211}, 2024. [Online]. Available: \url{https://arxiv.org/abs/2405.06211}
\BIBentrySTDinterwordspacing

\bibitem{toro2024dynamic}
S.~Toro, A.~V. Anagnostopoulos, S.~M. Bello, K.~Blumberg, R.~Cameron, L.~Carmody, A.~D. Diehl, D.~M. Dooley, W.~D. Duncan, P.~Fey \emph{et~al.}, ``Dynamic retrieval augmented generation of ontologies using artificial intelligence (dragon-ai),'' \emph{Journal of Biomedical Semantics}, vol.~15, no.~1, p.~19, 2024.

\bibitem{lopez2025clear}
\BIBentryALTinterwordspacing
I.~Lopez, A.~Swaminathan, K.~Vedula, S.~Narayanan, F.~Nateghi~Haredasht, S.~P. Ma, A.~S. Liang, S.~Tate, M.~Maddali, R.~J. Gallo, N.~H. Shah, and J.~H. Chen, ``Clinical entity augmented retrieval for clinical information extraction,'' \emph{npj Digital Medicine}, vol.~8, no.~1, p.~45, 2025. [Online]. Available: \url{https://www.nature.com/articles/s41746-024-01377-1}
\BIBentrySTDinterwordspacing

\bibitem{zhang2025longcontext}
\BIBentryALTinterwordspacing
G.~Zhang, Z.~Xu, Q.~Jin, F.~Chen, Y.~Fang, Y.~Liu, J.~F. Rousseau, Z.~Xu, Z.~Lu, C.~Weng, and Y.~Peng, ``Leveraging long context in retrieval augmented language models for medical question answering,'' \emph{npj Digital Medicine}, vol.~8, no.~1, p. 239, 2025. [Online]. Available: \url{https://www.nature.com/articles/s41746-025-01651-w}
\BIBentrySTDinterwordspacing

\bibitem{wu2024medgrphrag}
J.~Wu, J.~Zhu, Y.~Qi, J.~Chen, M.~Xu, F.~Menolascina, and V.~Grau, ``Medical graph rag: Towards safe medical large language model via graph retrieval-augmented generation,'' \emph{arXiv preprint arXiv:2408.04187}, 2024.

\bibitem{feng2025hyperrag}
\BIBentryALTinterwordspacing
Y.~Feng, H.~Hu, X.~Hou, S.~Liu, S.~Ying, S.~Du, H.~Hu, and Y.~Gao, ``{Hyper-RAG}: Combating {LLM} hallucinations using hypergraph-driven retrieval-augmented generation,'' \emph{arXiv preprint arXiv:2504.08758}, 2025. [Online]. Available: \url{https://arxiv.org/abs/2504.08758}
\BIBentrySTDinterwordspacing

\bibitem{allMiniLM_L6_v2}
{UKP Lab / Hugging Face / Sentence Transformers}, ``sentence-transformers / all-minilm-l6-v2,'' \url{https://huggingface.co/sentence-transformers/all-MiniLM-L6-v2}, accessed: 2025-10-06.

\bibitem{douze2024faiss}
M.~Douze, A.~Guzhva, C.~Deng, J.~Johnson, G.~Szilvasy, P.-E. Mazar{\'e}, M.~Lomeli, L.~Hosseini, and H.~J{\'e}gou, ``The faiss library,'' \emph{arXiv preprint arXiv:2401.08281}, 2024.

\bibitem{Whoosh}
M.~Chaput, ``Whoosh: A fast, pure-python full-text indexing and search library,'' \url{https://whoosh.readthedocs.io/}, 2016, version 2.7.4, BSD License; accessed 2025-10-06.

\bibitem{openai2023gpt4}
\BIBentryALTinterwordspacing
{OpenAI}, ``Gpt-4 technical report,'' 2023, arXiv preprint arXiv:2303.08774. [Online]. Available: \url{https://arxiv.org/abs/2303.08774}
\BIBentrySTDinterwordspacing

\bibitem{google2024gemini1_5}
\BIBentryALTinterwordspacing
{Google AI / DeepMind}, ``Gemini 1.5 pro: Next-generation multimodal model,'' 2024, model card / system documentation. [Online]. Available: \url{https://blog.google/technology/ai/google-gemini-next-generation-model-february-2024}
\BIBentrySTDinterwordspacing

\end{thebibliography}

\end{document}